\documentclass{article}

\usepackage[nonatbib, preprint]{neurips_2023}

\usepackage[utf8]{inputenc} 
\usepackage[T1]{fontenc}    
\usepackage{hyperref}       
\usepackage{url}            
\usepackage{booktabs}       
\usepackage{amsfonts}       
\usepackage{nicefrac}       
\usepackage{microtype}      
\usepackage{xcolor}         
\usepackage{dirtree}
\usepackage{bbding}
\usepackage{multirow}

\usepackage{graphicx}
\usepackage{framed}
\usepackage{tcolorbox}

\usepackage{url}
\usepackage{rotating}
\usepackage{subfigure}
\usepackage{booktabs}
\usepackage{amsmath,bm}
\usepackage{bbm}
\usepackage{cite}
\usepackage{enumerate}
\usepackage{algorithm}
\usepackage{algorithmic}
\usepackage{colortbl}
\usepackage{psfrag}
\usepackage{adjustbox}
\usepackage{wrapfig}
\usepackage{xspace} 
\usepackage{amssymb}
\usepackage{multirow}
\usepackage{afterpage}
\usepackage{float}

\newcommand{\answerTODO}[1][]{\textcolor{red}{\bf [TODO]}}

\usepackage{enumitem}

\usepackage{xcolor,colortbl}

\definecolor{Gray}{gray}{0.85}
\definecolor{LightCyan}{rgb}{0.88,1,1}

\newcolumntype{a}{>{\columncolor{Gray}}c}
\newcolumntype{b}{>{\columncolor{white}}c}

\usepackage{xcolor}
\colorlet{darkgreen}{green!65!black}
\colorlet{darkblue}{blue!75!black}
\colorlet{darkred}{red!80!black}
\definecolor{lightblue}{HTML}{0071bc}
\definecolor{lightgreen}{HTML}{39b54a}
\definecolor{manyshot}{HTML}{6969ff}
\definecolor{medshot}{HTML}{f7c600}
\definecolor{fewshot}{HTML}{ff6969}
\definecolor{mypurple}{HTML}{412F8A}
\definecolor{myorange}{HTML}{fc8e62}
\definecolor{deemph}{gray}{0.55}

\definecolor{linkcolor}{HTML}{ED1C24}

\usepackage[colorinlistoftodos]{todonotes}

\newcommand{\gtext}[1]{\textcolor{green!50!black}{#1}}

\renewcommand{\paragraph}[1]{\vspace{1.25mm}\noindent\textbf{#1}}

\definecolor{baselinecolor}{gray}{.95}

\title{Large Language Models are Few-Shot \\ Health Learners}

\author{%
  Xin Liu, Daniel McDuff, Geza Kovacs, Isaac Galatzer-Levy, Jacob Sunshine, \\ 
  \textbf{Jiening Zhan, Ming-Zher Poh, Shun Liao, Paolo Di Achille, Shwetak Patel} \\
  Consumer Health Research Team\\
  Google\\
  \texttt{\{xliucs,dmcduff\}@google.com} \\
}

\begin{document}

\maketitle

\begin{abstract}

Large language models (LLMs) can capture rich representations of concepts that are useful for real-world tasks. However, language alone is limited. While existing LLMs excel at text-based inferences, health applications require that models be grounded in numerical data (e.g., vital signs, laboratory values in clinical domains; steps, movement in the wellness domain) that is not easily or readily expressed as text in existing training corpus. We demonstrate that with only few-shot tuning, a large language model is capable of grounding various physiological and behavioral time-series data and making meaningful inferences on numerous health tasks for both clinical and wellness contexts. Using data from wearable and medical sensor recordings, we evaluate these capabilities on the tasks of cardiac signal analysis, physical activity recognition, metabolic calculation (e.g., calories burned), and estimation of stress reports and mental health screeners.
\end{abstract}

\vspace{-0.5cm}
\section{Introduction}

The fundamental ability of large language models (LLMs) to capture knowledge and concepts has improved dramatically as neural architectures have been scaled. Exponentially greater numbers of parameters and training samples has led to models that can encode vast amounts of information~\cite{brown2020language,chowdhery2022palm,thoppilan2022lamda,openai2023gpt4,touvron2023llama}. 
These models present opportunities in many domains including software engineering~\cite{chen2021evaluating}, information retrieval and data mining~\cite{gu2021domain}, content creation~\cite{chung2022talebrush} and health~\cite{singhal2022large}. However, in many cases text alone is not enough. Grounding in non-linguistic observations such as actions, experiences or perceptions is an essential part of deriving meaning from language~\cite{glenberg2005grounding} as words alone does not capture all the necessary information to complete a given task. Tuning, or grounding, of this kind is necessary to connect the knowledge contained within a language model to events or actions in the physical world~\cite{ahn2022can}. For these reasons, models have been combined with data from other modalities, most commonly images~\cite{mostafazadeh2017image,huber2018emotional,li2022grounded}, audio~\cite{xu2021text}, video~\cite{zhou2019grounded,soldan2022mad} or physical actions~\cite{ahn2022can}.

Despite the importance of physiology and behavior to human health, evaluating and grounding models with these data has been relatively unexplored. Medical-domain language models have shown remarkable capabilities to capture sophisticated knowledge~\cite{gu2021domain,singhal2022large}, performing well on multiple-choice examination questions (e.g., board exams). However, they have not been tested extensively on tasks that involve analysis of physiological (e.g., heart rate) and behavioral timeseries (e.g., daily steps), which are important for many consumer health tasks. Without data of this kind, derived from actual measurements of the human body, these models may be limited in their ability to provide specific health insights, particularly when the insight involves inference from physiological data. For example, an individual searching for a diagnosis for their ailment may have limited success if the model has no insight into physiology (e.g., elevated heart rate, fever) that could be underlying their symptoms. Conversely, models capable of incorporating user physiological data may be able to provide enhanced insight to queries within a medical domain. We hypothesize that LLMs have some capability to interpret and reason about health information, but that that ability will dramatically improve with grounding from even a small number of examples of numerical health data.

Physiological sensing and monitoring represents the quantitative measurement of physiological processes in the human body and is a cornerstone of clinical-based diagnosis and patient management. Such monitoring is now also commonplace using commodity consumer electronics such wearables, phones and other low-cost sensing platforms. 
Wearable devices and unobtrusive sensing have helped individuals realize meaningful changes in their health, such as helping to increase the amount of physical activity people engage in~\cite{ferguson2022effectiveness}. Moreover, when done thoughtfully and in an evidence-based manner, it is generally accepted that helping individuals derive insight from their data could increase the frequency of engaging in beneficial health behaviors.  Traditionally, researchers have had to collect substantial amounts of individual health data for each task (e.g., to train an atrial fibrillation classifier \cite{lubitz2022detection}). This process is time-consuming and resource-intensive. Yet, there is evidence that LLMs have the potential to act as universal pretrained models and perform zero- or few-shot inferences (e.g., translation, question-answering) \cite{reiss2012introducing}. 
However, it is not clear to what extent this is true for tasks that involve physiological or behavioral data. For example, classifying atrial fibrillation from a series of interbeat-intervals not only requires understanding of how to process time-series data but also domain knowledge in cardiology. Can LLMs act as efficient few-shot learners of arrhythmia?

In this paper, we compile a dataset of common consumer health tasks (e.g. activity recognition, heart rhythm classification) to explore whether LLMs can ground time-series health data and serve as universal few-shot learners. We systematically demonstrate that with few-shot prompt tuning LLMs can ground numerical time series data, derived from wearable and clinical-grade sensing devices, resulting in large improvements over zero-shot inferences and supervised baselines on tasks as varied as activity recognition, computing calories burned and atrial fibrillation classification. 
To summarize, our contributions are to: 
1) curate a dataset of numerical consumer health tasks for evaluating LLM performance in grounding time-series data.
2) demonstrate the capability of a large language model as a universal few-shot learner for a range of wellness and clinical level health tasks. 
3) expose the limitations of pretrained large language models on these tasks.
4) present an set of physiologically and behaviorally tuned models that perform efficiently with a small number of training examples.
To the best of our knowledge, this is the first paper exploring grounding LLMs with time-series physiological and behavioral data and using LLMs as few-shot learners on health tasks.

\begin{figure*}[t!]
  \centering
    \includegraphics[width=\textwidth]{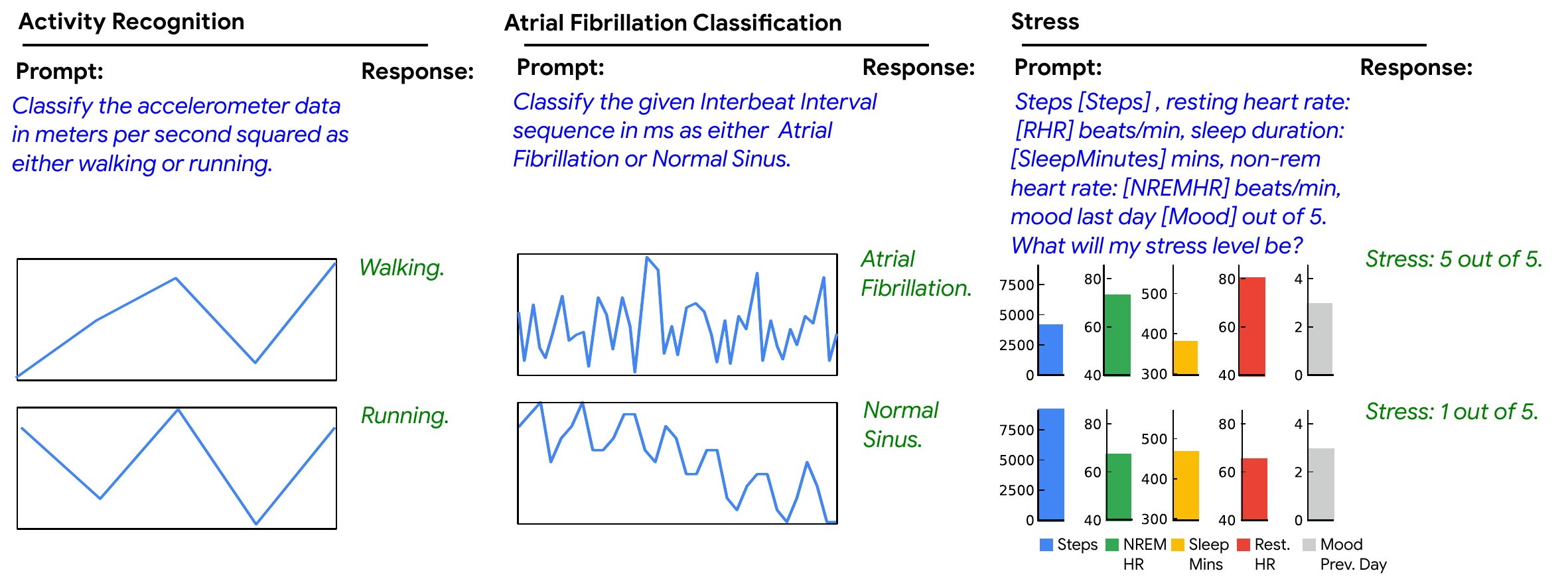}
    \caption{\textbf{Examples of question-answer pairs for our health tasks.} In the prompts, data were represented numerically rather than graphically.}
      \label{fig:data_examples}
     \vspace{-0.8cm}
  \end{figure*}

\section{Background}

\textbf{Large Language Models.}
The ability for language models trained with vast corpora of text data to produce text that is indistinguishable from that written by humans~\cite{openai2023gpt4} and to reach human performance on multiple benchmarks~\cite{wang2019superglue} is impressive. It is possible for these models to encode a large amount of knowledge from a range of domains. One of the most significant properties of LLMs is their capability to perform zero-shot or few-shot inferences, a product of the large amount of text from diverse domains that was used to train them~\cite{kaplan2020scaling,touvron2023llama}. These LLMs have so far obeyed scaling laws with consistent, smooth model improvement as the model, dataset and compute size used for training is a compelling argument for further improvements~\cite{kaplan2020scaling}.

While general domain models can encode medical data, models fine-tuned specifically on medical literature have been produced that improve upon these benchmarks~\cite{gu2021domain,singhal2022large}. Passing the board exam, an examination typically taken by second year medical students in the United States, has been touted as a milestone in model performance.  These results are impressive and surprising, but the models do have weaknesses. Inaccuracies and hallucination of data, poor ``awareness'' about their abilities, and generation of responses that are unduly confident or authoritative, are all reasons that they present dangers. These dangers are particularly problematic for safety critical tasks in health and medicine. Proper grounding, tuning and evaluation are all necessary, as training on a large amount of data cannot be relied on to produce good results on every task.

\textbf{Arithmetic.} The ability of language models to solve arithmetic tasks is one such example. Research into the arithmetic capabilities of language models has often limited investigations to relatively simple operations (addition, subtraction, multiplication, division, log transformations) as performance on more complicated tasks is poor. Evidence shows that larger parameter models trained on more data do perform better at these operations than smaller models trained on less data~\cite{yuan2023well}. However, as the tasks become more complex and input sequences are increased in length their performance drops~\cite{qian2022limitations}.

\textbf{Health.} Wearable and cardiovascular health sensing is important for monitoring physical and behavioral states. Aging populations, the increasing burden of chronic diseases and disruptive global events (e.g., the COVID-19 pandemic) have raised the urgency for health technologies that support remote care and wellbeing. However, raw sensor data is often difficult for people and LLMs to consume. Processing  classifying types of workouts and activities, computing calories burned, detecting arrhythmia and triggering interventions to stress responses are all examples of these.

\begin{table}[!t]
\setlength{\tabcolsep}{3pt}
\caption{\small 
\textbf{Consumer health tasks.} We compile a dataset encompassing nine tasks, across four domains. 
}
\vspace{-8pt}
\label{tab:tasks}
\small
\begin{center}
\adjustbox{max width=\textwidth}{
\begin{tabular}{cllp{5cm}l}
\toprule[1.5pt]
\textbf{Topic} & \textbf{Data} & \textbf{Task} & \textbf{Prompt} & \textbf{Target} \\ \hline 

\parbox[t]{2mm}{\multirow{13}{*}{\rotatebox[origin=c]{90}{Cardiovascular}}} & \multirow{13}{*}{MIT-BIH \& MIMIC-III \cite{johnson2016mimic, goldberger2000physiobank}} &
Instantaneous HR to Average HR  & Given a sequence of heart rates, calculate the average heart rate: \textcolor{darkgreen}{\texttt{[HR]}} &      \textcolor{darkblue}{ $\mu$(\texttt{[HR])} } \\

 & & IBIs to HR & Calculate heart rate based on the given inter-beat intervals in ms: \textcolor{darkgreen}{\texttt{[IBI]}} & 
\textcolor{darkblue}{$\frac{60}{ \mu(\texttt{[IBI]})}$} \\

 
 & & IBIs to Atrial Fibrillation & Classify the given Interbeat Interval sequence in ms as either Atrial Fibrillation or Normal Sinus: \textcolor{darkgreen}{\texttt{[IBI]}} & \textcolor{darkblue}{\texttt{[A.Fib.]}} \\
 
 & & IBIs to Sinus Bradycardia  & Classify the given Interbeat Interval sequence in ms as either Sinus Bradycardia or Normal Sinus: \textcolor{darkgreen}{\texttt{[IBI]}} & \textcolor{darkblue}{\texttt{[Sinus. B.]}} \\
  
  &  & IBIs to Sinus Tachycardia  & Classify the given Interbeat Interval sequence in ms as either Sinus Tachycardia or Normal Sinus: \textcolor{darkgreen}{\texttt{[IBI]}} & \textcolor{darkblue}{\texttt{[Sinus T.]}} \\

 \midrule
\parbox[t]{2mm}{\multirow{3}{*}{\rotatebox[origin=c]{90}{Activity}}}  & \multirow{3}{*}{PAMAP2~\cite{reiss2012introducing}} & Activity Recognition & Classify the following accelerometer data in meters per second squared as either walking or running: \textcolor{darkgreen}{\texttt{[Acc]}} & \textcolor{darkblue}{\texttt{[Activity]}}  \\
   \midrule
   
\parbox[t]{2mm}{\multirow{4}{*}{\rotatebox[origin=c]{90}{Metabolic}}} & \multirow{4}{*}{Synthetic} 
 & Calorie Estimate  & How many total calories will I burn after \textcolor{darkgreen}{\texttt{[Activity]}} for \textcolor{darkgreen}{\texttt{[Duration]}}  minutes? My weight is \textcolor{darkgreen}{\texttt{[Weight]}} lbs. & \textcolor{darkblue}{\texttt{[Calories]}} \\
 \\\midrule
 
\parbox[t]{2mm}{\multirow{12}{*}{\rotatebox[origin=c]{90}{MHealth}}} & \multirow{12}{*}{Fitbit}
& Estimation of Daily Stress EMA  & Steps: \textcolor{darkgreen}{\texttt{[Steps]}}, resting heart rate: \textcolor{darkgreen}{\texttt{[RHR]}} beats/min, sleep duration: \textcolor{darkgreen}{\texttt{[SleepMinutes]}} minutes, non-REM heart rate \textcolor{darkgreen}{\texttt{[NREMHR]}} beats/min, mood last day \textcolor{darkgreen}{\texttt{[Mood]}} out of 5. What will my stress level be? & \textcolor{darkblue}{\texttt{[Stress]}}  \\
 & & Estimation of PHQ8 Score & Steps: \textcolor{darkgreen}{\texttt{[Steps]}}, resting heart rate: \textcolor{darkgreen}{\texttt{[RHR]}} beats/min, sleep duration: \textcolor{darkgreen}{\texttt{[SleepMinutes]}} minutes, non-REM heart rate \textcolor{darkgreen}{\texttt{[NREMHR]}} beats/min. feeling last month: \textcolor{darkgreen}{\texttt{[Mood]}} out of 5. What will my PHQ score be?  & \textcolor{darkblue}{\texttt{[PHQ]}} \\
\bottomrule[1.5pt]
\end{tabular}}
\footnotesize
IBI = Interbeat Interval, HR = Heart Rate, RHR = Resting HR, NREMHR = Non-REM HR, PHQ = Patient Health Questionnaire, EMA = Ecological Momentary Assessment
\end{center}
\vspace{-0.8cm}
\end{table}

\section{Health Tasks}

In order to evaluate language model performance on numerical health problems, we choose a set of tasks with varying degrees of complexity (see Table~\ref{tab:tasks} for a summary). Simple tasks (i.e., averaging instantaneous heart measurements) involve trivial mathematical operations (i.e., addition and division), whereas complex tasks (i.e., predicting an individual's score on a mental health screening questionnaire) involve interpreting the context of numerical data and how they relate to mental states (i.e., what the step count represents in terms of an individual's physical activity and how their amount of physical activity might relate to their mental health). Below we describe all the tasks and the data used to evaluate them (for additional details please see the supplementary material).

\textbf{Cardiovascular.} Electrocardiograms (ECGs) and photoplethysmograms (PPGs) serve as critical tools in understanding the functioning of the heart. These waveforms can be used to detect heart beats, from which interbeat intervals (IBIs) can be computed. Specifically, we choose five cardiovascular tasks involving interpreting HR and cardiac IBIs: 1) computation of average heart rate (HR) from a series of instantaneous HR measurements, 2) calculation of average heart rate (in beats/min) from interbeat intervals (in milliseconds), 3) atrial fibrillation classification, 4) sinus bradycardia classification, and 5) sinus tachycardia classification. Tasks 3-5 are all framed as binary classification. Our cardiovascular tasks encompass both classification and regression learning tasks. We aspire for our language model to utilize its knowledge (for instance, recognizing specific patterns associated with atrial fibrillation) to enhance the accuracy and effectiveness of its predictions. For our cardiovascular tasks, we draw upon the MIT-BIH Atrial Fibrillation Database~\cite{goldberger2000physiobank} and the MIMIC-III dataset~\cite{johnson2016mimic}. These data are collected from IRB approved studies with informed consent from all the research subjects. From these datasets, we sample 25 samples of each class for training and 100 samples for testing. The details of data split and experiments are included in supplementary materials.  

\textbf{Metabolic.} Assessing energy expenditure is an essential aspect of health and wellness monitoring. The estimation of calories burned based on physical activity is a complex calculation that requires the incorporation of multiple factors. In our study, we leverage LLMs to estimate caloric expenditure based on the metabolic equivalent of task (MET) formula: 

\begin{tcolorbox}[boxsep=1pt, left=4pt, right=4pt, top=4pt, bottom=4pt, box align=center, halign=center]
$\textit{Calories} = \textit{MET Value} \times \textit{Duration} \times \textit{Weight} \div 200$
\end{tcolorbox}

To facilitate this, we construct a synthetic dataset combining an individual's weight, their physical activity characterized by MET values (8.5 for running, 7.5 for biking, and 3.5 for walking), and the duration of those activities. Our task involves calculating calories burned using only activity type, duration, and weight. This task requires LLMs to accurately recall the MET value for each activity and the equation of how to calculate burned calories. As such, the task highlights the model's potential in both regression (estimating continuous calories burned) and numerical reasoning capabilities. The synthetic metabolic health data includes 25 samples for training and 100 samples for testing.

\textbf{Activity Recognition.} Recognizing physical activities constitutes a vital component of health and wellness monitoring. In this paper, we focus on the classification of two fundamental activities: walking and running. We employ the PAMAP2 Physical Activity Monitoring Dataset \cite{reiss2012introducing}, which offers a broad spectrum of physical activity data. We select the walking and running classes for the purpose of this study as multi-class classification with a very large number of classes was poor. We downsample the data to second-level granularity by taking the mean magnitude of non-overlapping 1-second windows. This metric captures the magnitude and intensity of the activities in a reasonable fashion. For each task, a 5-second window of these features is added to the prompt. The models are then tasked to classify the activity as either walking or running. This task requires the model to leverage internal knowledge about the characteristics of accelerometer patterns associated with walking and running and perform a classification of the time-series data. We randomly sample balanced 25 samples of each class for training and 100 samples for testing.

\textbf{Mental Health.} Assessment of psychological functioning and psychiatric symptoms are linguistic in nature, measured through semi-structured interviews and self report. As a result, the use of language models is heavily researched~\cite{berger2021using}, demonstrating success in instances like the classification of depression and other psychiatric disorders~\cite{zhang2022natural}, but typically lacking the training data necessary for robust generalized models. Large language models may have the potential for success in accurate psychiatric classification because of the large corpus of data that they are trained on. Many of the symptoms are behavioral or physiological descriptors that can be accurately measured through sensors, though they are currently assessed conversationally. As an example, core symptoms of depression and anxiety include changes in sleep and movement patterns. 
In this paper, we take on two tasks including 1) prediction of daily ecological momentary stress prediction based on daily wearable behavior and physiology 2) depression score/classification prediction based on the patient health questionnaire (PHQ-8~\cite{kroenke2009phq}) and wearable behavior and  physiology over a two week period, consistent with the assessment period of the PHQ-8. 

In this work, we leverage a large corpus of consented data from Fitbit devices. These trackers and smartwatches have been used in a large number of validation studies and clinical trials and advances in device quality and capabilities offer new opportunities for research. Specifically, we use data from a longitudinal study to analyze the relationship between wearable physiological and behavioral data, and self-reported measures of mental health and well-being \footnote{Reference removed because of anonymization.}. The data are collected with informed consent from all participants who opted-in to an Institutional Review Board (IRB) approved study (Institution removed for anonymization). During the study, the participants completed intake and outtake surveys, daily ecological assessments and enabled sensing of several features from their Fitbit devices. We use daily Fitbit metrics (step count, resting heart rate and heart rate during non-REM sleep and sleep duration) to predict daily stress EMAs (stressed vs. not stressed) and four week average metrics of the same variables to predict PHQ-8 scores at the end of the four week period. In this work we randomly sample data from 50 different participants to form the training set and 100 participants to form the testing set. For stress prediction a single day is sampled from each participant to give 50 and 100 days of Fitbit data with a stress EMA in the training and testing set respectively.  For the PHQ prediction the Fitbit data for each participant is averaged across the study duration (4-weeks) as features and associated with their PHQ score at the end of the study to give 50 (25 for each class) and 100 samples in the training and testing set respectively. More details are included in supplementary materials.

\vspace{-0.3cm}
\section{Method \& Experiment}

\subsection{Architecture}

We implement our experiments on a 24 billion parameter transformer architecture (PaLM~\cite{chowdhery2022palm}).  
Pretrained on a massive text corpus comprising 780 billion tokens, the model has been exposed to a diverse set of natural language use cases drawn from various sources, including filtered webpages, books, Wikipedia articles, news stories, social media conversations, and even code. The pretraining dataset features not only language data but also programming code, the latter of which was sourced from open repositories on GitHub. The resulting pretrained model has significant language and reasoning capabilities. Nevertheless, as we will demonstrate, it is not inherently capable of producing dependable answers to inquiries that pertain to physiological and behavioral data.

\subsection{Physiological Grounding for Few-Shot Learning}

\begin{figure*}[t!]
\centering
  \includegraphics[width=\textwidth]{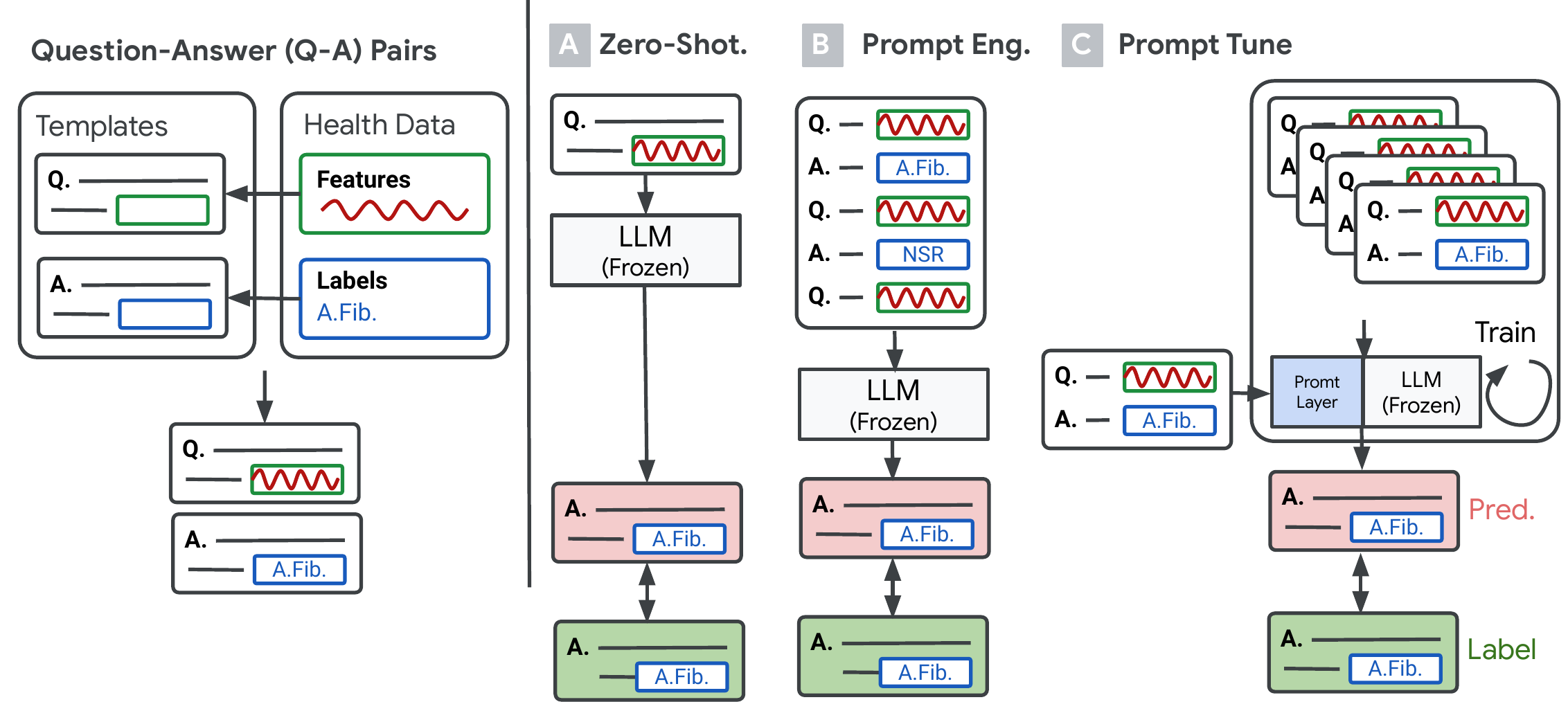}
  \caption{\textbf{Our dataset construction and training configurations.} We use physiological data to construct question-answer pairs. We create training, validation and test splits and compare zero-shot prediction performance to models created via prompt engineering and prompt tuning.}
    \label{fig:training}
\end{figure*}

For each task, we embed quantitative numerical data into textual templates (see Fig.~\ref{fig:training}) to create question answer pairs. Examples of these prompts and the numerical data are shown in Fig.~\ref{fig:data_examples} and Table \ref{tab:tasks}. We perform three different types of experiments with the language model: zero-shot evaluation, and prompt engineering and prompt tuning followed by evaluation.

\textbf{Zero-Shot.} In order to evaluate the zero-shot performance, we conduct tests on all tasks without fine-tuning the model weights or providing any examples to the model. Specifically, we utilize textual templates and input numerical physiological data, such as a sequence of inter-beat intervals or high-level sensor data (e.g., heart rate or sleep hours), along with specific context and questions related to the target task. For instance, the zero-shot prompt for the atrial fibrillation (A.Fib.) classification task is as follows:

\begin{framed}
\centering
\textit{"Classify the given interbeat interval sequence in ms as either Atrial Fibrillation or Normal Sinus: \gtext{ 896,1192,592,1024,1072,808,888,896,760,1000,784,736,856,1000,1272,824,\\872,1120,840,896,888,560,1248,824,968,960,1000,1008,776,744,896,1256.}"}
\end{framed}

\textbf{Prompt Tuning.} In order to incorporate numerical time-series physiological and behavioral data into the pretrained language model, we add a soft learnable prompt embedding \cite{lester2021power}. This layer is responsible for learning a task-specific prompt embedding that is attached to each sample during inference with the frozen language model, in preparation for downstream tasks. This approach enables us to move beyond manual prompt design and facilitates learning of an optimal prefix for each task. In our particular case, the prompt layer encodes information that assists the language model in comprehending numerical and time-series data that is absent in the original frozen language model. Specifically, the size of the prompt embedding is $4096 \times 1$. 
The input for prompt tuning is similar to that used in the zero-shot setting, the only difference is that prompt tuning requires paired labels for each individual task. For instance, the input and the corresponding label for the activity recognition task would be: 

\begin{framed}
\centering
Input: \textit{"Classify the following accelerometer data in meters per second squared as either walking or running: \gtext{0.052,0.052,0.052,0.051,0.052,0.055,0.051,0.056,0.06,0.064}"} \\
Label: "Running"
\end{framed}

To better understand how language can utilize inherent domain knowledge, we present two unique sets of prompts that are used during prompt tuning for cardiovascular and activity recognition tasks: 1) \emph{Context-Inclusive} Prompts, and 2) \emph{Numerical-Only} Prompts. 

The \emph{Numerical-Only} prompt approach presents only the sequence of numbers to the model, devoid of any additional textual context. This strategy aims to test the language model's ability to discern patterns and generate accurate predictions based purely on the input numerical data, without any supplementary information about the nature and source of the data. We elected not to use the Numerical-Only prompt for tasks associated with mental health and metabolic calorie expenditure. As these tasks do not involve sequential data, and it is difficult to imagine how context-free numerical data would yield any meaningful prediction. Furthermore, in these situations, the contextual cues are particularly vital for the model to comprehend the task and apply its domain knowledge.

The \emph{Context-Inclusive} prompts involve incorporating relevant contextual information into the prompt. For instance, in cardiovascular tasks, we unambiguously indicated that the input sequence signifies Inter-beat Intervals (IBIs). Similarly, in the activity recognition task, we clarify that the input data originates from accelerometer readings and provided the units of measurement. The inclusion of this additional context is based on the hypothesis that it could steer the language model towards utilizing its domain-specific knowledge, potentially improving its predictive accuracy.

In summary, these two prompt tuning strategies enable us to investigate the influence of contextual information on the model's performance. Significantly, they offer insights into the model's capacity to harness its domain knowledge when provided with suitable hints, as well as its ability to identify patterns and execute tasks without explicit contextual information. We assess our prompt tuning method in 3-shot, 10-shot, and 25-shot scenarios. The language model, augmented with an additional tunable prompt layer, is trained over 5000 steps using a learning rate of 3.0.

\textbf{Prompt Engineering.} As with zero-shot learning, in this setting we leverage a pre-trained model to perform inference with frozen weights. However, we incorporate additional context within the input prompt as a series of examples before asking the test question. The examples guide the language model toward the structure of the task. As an example, in the calorie estimation task a prompt might be framed as follows:

\begin{framed}
\centering
\textit{"Question: How many total calories will I burn after \gtext{walking} for \gtext{50} minutes? My weight is \gtext{156 lbs}. Answer: 214.
\\ Question: How many total calories I bured after \gtext{walking} for \gtext{48} minutes? My weight is \gtext{221} lbs. Answer: "}
\end{framed}

In this paper, we specifically focus on a 3-shot setting for prompt engineering, largely due to the nature of our data. Given that our data consisted of sequential and time-series information, the samples are typically lengthy. However, LLMs are subject to a limit on the number of tokens that can be inputted, which motivated our decision to concentrate on a 3-shot setting. 
It's noteworthy that this prompt engineering approach did not involve any additional training of the model. Instead, it relied entirely on the out-of-the-box capabilities of the pre-trained model, testing its ability to perform tasks based on the provided prompts and its existing knowledge. This experiment provide valuable insights into the model's few-shot performance capabilities and the effectiveness of carefully engineered prompts in guiding the model towards desired task performance compared against to prompt tuning. The results of prompt engineering are included in the supplementary materials. 

\textbf{Supervised Baseline.} Finally, we juxtapose all these grounding methods with a supervised learning baseline for comparison. In this paper, we intentionally maintain a small number of training samples in order to evaluate few-shot performance. Nevertheless, we find it valuable to establish a supervised baseline, as it helps discern whether large language models could offer additional benefits by leveraging the knowledge encapsulated within pre-trained weights, knowledge that the training samples themselves might not encapsulate. To this end, we utilize a multilayer perceptron (MLP) with $4096 \times 1$, training it on the same samples used for prompt tuning in our large language model.

We use 32 TPU-v4 chips for all our experiments including training and evaluation.

\begin{table}[t]
\setlength{\tabcolsep}{3pt}
\caption{\small \textbf{Results.} Comparison of performance between prompt-tuned LLMs (w/ Context-Inclusive Prompts) and supervised neural network training across all consumer health tasks.}
\vspace{-8pt}
\label{exp:table:finetune}
\small
\begin{center}
\adjustbox{max width=\textwidth}{
\begin{tabular}{llccccccca}
\toprule[1.5pt]
& & & \multicolumn{3}{c}{\textbf{Supervised Baseline}} & \multicolumn{3}{c}{\textbf{LLM with Context}} \\
\cmidrule(lr){4-6} \cmidrule(lr){7-9} 
\textbf{Topic} & \textbf{Task}  & \textbf{Metric} & 3-Shot & 10-Shot & 25-Shot & 3-Shot & 10-Shot & 25-Shot & \% Improvement 
\\ \hline \hline

\multirow{3}{*}{Cardio} & HRs to Average HR & MAE $\downarrow$ (beats/min) & 3.41 & 1.37 & 1.08 & 6.00 & 2.49 & \textbf{1.06} & \textcolor{darkgreen}{\texttt{+}\textbf{1.90\%}}\\
& IBIs to HR & MAE $\downarrow$ (beats/min) & 34.0 & 20.0 & 19.8 & 12.3 & 5.87 & \textbf{5.01} & \textcolor{darkgreen}{\texttt{+}\textbf{74.7\%}}\\
& IBIs to A.Fib. & Accuracy $\uparrow$ (\%)  & 52.5 & 72.5 & 75.0 & 85.0 & 75.0 & \textbf{89.0} & \textcolor{darkgreen}{\texttt{+}\textbf{19.7\%}}\\ 
& IBIs to Sinus B. & Accuracy $\uparrow$ (\%)  & 88.0 & 86.0 & 86.0 & 81.0 & 79.0 & \textbf{92.0} & \textcolor{darkgreen}{\texttt{+}\textbf{7.00\%}}\\ 
& IBIs to Sinus T. & Accuracy $\uparrow$ (\%)  & 56.0 & 53.0 & 61.0 & 65.0 & 82.0 & \textbf{88.0} & \textcolor{darkgreen}{\texttt{+}\textbf{44.3\%}}\\ \hline

\multirow{1}{*}{Activity} & IMU Activity & Accuracy $\uparrow$ (\%) & 56.0 & 60.0 & 64.0 & 62.0 & 80.0 & \textbf{85.0} & \textcolor{darkgreen}{\texttt{+}\textbf{32.8\%}}\\ \hline

\multirow{1}{*}{Metabolic} & Calories & MAE $\downarrow$ (calories) & 185 & 97 & 89 & 106 & 77 & \textbf{48} & \textcolor{darkgreen}{\texttt{+}\textbf{46.1\%}}\\ \hline

\multirow{2}{*}{MHealth} & Fitbit to Stress & Accuracy $\uparrow$ (\%)  & 37.5 & 70.5 & 80.0 & 72.5 & 71.5 & \textbf{82.5} & \textcolor{darkgreen}{\texttt{+}\textbf{3.10\%}}\\
& Fitbit to PHQ & Accuracy $\uparrow$ (\%)  & 51.0 & 52.0 & 53.0 & 49.0 & 59.0 & \textbf{69.0} & \textcolor{darkgreen}{\texttt{+}\textbf{30.2\%}}\\

\bottomrule[1.5pt]
\end{tabular}}
\end{center}
\vspace{-0.2cm}
\end{table}

\section{Result \& Discussion}

\begin{figure*}
  \centering
    \includegraphics[width=\textwidth]{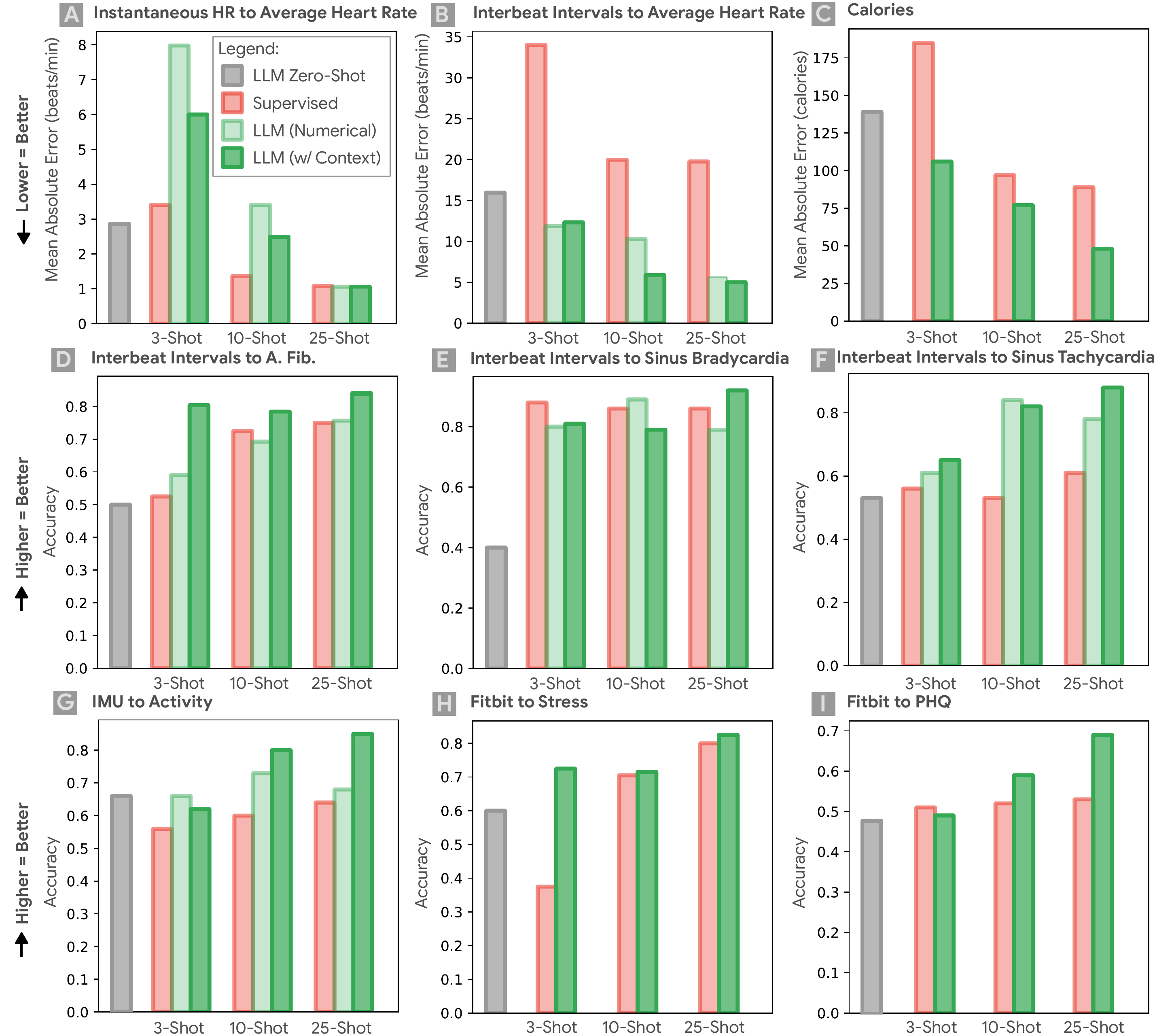}
    \caption{
     \textbf{Results of 3, 10 and 25-shot experiments for nine health tasks.}. Prompts include IBI sequences (in milliseconds) and the model is asked to provide instantaneous heart rate (A), average heart rate in beats per minute (B), presence or absence of atrial fibrillation (D), presence or absence of slowed heart rate/bradycardia (E) and elevated heart rate/Tachycardia (F). Accelerometer data into the prompt and the model is asked to classify walking or running (G); Fitbit data (e.g., steps, sleep hours) are included in the prompt and the model is asked to classify stress (H) and PHQ score (I). Data on exercise type, duration and weight are included in the prompt to estimate calories burned (C). The LLM with context-inclusive prompts outperforms the supervised baseline by up to 75\%.
    } 
      \label{fig:results}
  \end{figure*}

\textbf{Are language models effective few-shot learners for health tasks?} As illustrated in Table \ref{exp:table:finetune} and Figure \ref{fig:results}, the LLM functions as a universal few-shot learner for various consumer health tasks. These tasks assume a number of operations: interpreting time-series/sequential data, executing arithmetic operations, understanding context with domain knowledge, and recognizing patterns. We noted large decreases in error and increases in accuracy when the model was prompt tuned using only a small number of examples (from 3-shot to 10-shot and to 25-shot). A substantial benefit is observed when comparing our few-shot prompt tuning results with few-shot supervised training. For example, for five cardiovascular tasks, the 25-shot prompt tuned LLM performance improvement is up to 130 \% and the up to 75 \% when compared to zero-shot performance and supervised 25-shot baseline performance. Furthermore, we saw a 29\% increase in accuracy over zero-shot and a 33\% accuracy improvement in activity recognition tasks; a 66\% error reduction over zero-shot and a 46\% error reduction in calorie estimation; and up to a 44\% accuracy improvement over zero-shot and up to a 30\% accuracy improvement over the 25-shot supervised baseline in the mental health tasks using Fitbit data. These findings suggest that the capabilities of LLMs extend beyond processing textual data and traditional natural language processing tasks, demonstrating their versatility and potential for a wider range of consumer health applications.

\begin{table}[!t]
\setlength{\tabcolsep}{3pt}
\caption{\small \textbf{Numerical-Only vs Context-Inclusive.} Performance on tasks when the input prompt contains solely numerical values (Numerical-Only Prompts) versus when it includes textual context (Context-Inclusive Prompts).}
\vspace{-8pt}
\label{tab:hints}
\small
\begin{center}
\adjustbox{max width=\textwidth}{
\begin{tabular}{clccca}
\toprule[1.5pt]
& & & \multicolumn{3}{c}{\textbf{Performance}}  \\  \cmidrule(lr){4-6}
\textbf{Topic} & \textbf{Task} & \textbf{Metric} & No Context & Context & \% Improvement  \\ \hline \hline

\multirow{1}{*}{Cardio} 
  & HRs to Average HR & MAE $\downarrow$ (beats/min) & 1.06 & 1.06 & {\textbf{0\%}} \\ 
 & IBIs to HR & MAE $\downarrow$ (beats/min) & 5.63 & 5.01 & \textcolor{darkgreen}{\texttt{+}\textbf{11.0\%}}  \\ 
 & IBIs to A.Fib. & Accuracy $\uparrow$ (\%) & 75.6 & 89.0 & \textcolor{darkgreen}{\texttt{+}\textbf{17.7\%}} \\ 
 & IBIs to Sinus B. & Accuracy $\uparrow$ (\%) & 79.0 & 92.0 & \textcolor{darkgreen}{\texttt{+}\textbf{16.5\%}} \\ 
 & IBIs to Sinus T. & Accuracy $\uparrow$ (\%) & 78.0 & 88.0 & \textcolor{darkgreen}{\texttt{+}\textbf{12.8\%}} \\ \hline

\multirow{1}{*}{Activity} 
  & IMU to Activity & Accuracy $\uparrow$ (\%) & 68.0 & 85.0 & \textcolor{darkgreen}{\texttt{+}\textbf{25.0\%}}\\

\bottomrule[1.5pt]
\end{tabular}}
\end{center}
\vspace{-0.6cm}
\end{table}

\textbf{Are language models capable of zero-shot inferences?} Our findings indicate that a pretrained language model without explicit grounding may not necessarily perform well on a myriad of physiological and behavioral analysis tasks (see the gray bars in Figure \ref{fig:results}). Performance falls significantly short of the standard required for practical applications, for instance, an error of 139 in calorie estimation, and an accuracy score of 0.5 in atrial fibrillation classification are not better than could be achieved with a relatively naive model that guesses values. These results align with our expectations as language models are known to struggle with arithmetic tasks~\cite{qian2022limitations}. Given the multifaceted nature of our tasks, they may not be adequately represented in the training set. Therefore, we believe our findings are reasonable and expected. The details are included in supplementary materials. 

\textbf{Does offering more context to the LLM enhance performance?} Language models have the capacity to encode knowledge garnered from the language corpora used for their training. In addition to arithmetic proficiency, the tasks we evaluated necessitate domain-specific knowledge to interpret consumer health data accurately. Our results indicate that integrating more textual context within the question prompt enhances performance (see Table~\ref{tab:hints}) across all tasks, barring the very simple task of averaging HRs. This suggests that by supplementing the prompt with this context, we essentially anchor the input data within a specific domain. Anchoring helps the model to navigate its vast knowledge base and tap into the relevant information more effectively.

\textbf{What are the implications for consumer health research?} 
Our results and findings demonstrate the versatility and efficacy of LLMs in various consumer health tasks, assuming they can be appropriately tuned, and signify an opportunity for these models to play a substantial role in the future of health research and applications. They have shown promise in handling an array of computational tasks across different types of physiological and behavioral data. 
This implies that LLMs can be instrumental in dealing with complex health data, improving predictive accuracy, and possibly enhancing patient outcomes. Our exploration of prompt tuning strategies has unveiled the significant impact of contextual information on the model's performance. The increased performance of the model when presented with contextually rich prompts underscores the value of domain-specific information in driving accurate predictions. This insight could lead to more effective methodologies for training and utilizing AI models in the field of health research. Lastly, our findings highlight the remarkable capability of LLMs to function as effective few-shot learners, even with very limited examples. This opens up possibilities for their application in niche areas of consumer health research where extensive training data may not be readily available. In sum, the outcomes of our research point to a promising future where LLMs can contribute significantly to consumer health research, driving advancements in personalized healthcare, predictive modeling, and patient monitoring.


\textbf{What are the limitations and broader impacts of this work?}
In this paper, we only evaluated one LLM. Clearly differences exist across LLM capabilities; and many models struggle with arithmetic tasks. However, it is difficult to extrapolate exact performance to other models. We were careful to curate testing sets that contain a balanced distribution of samples; however, these sets were small.
Technologies that not only support healthcare, but also promote wellness, are very valuable to society as a whole. However, leveraging the rapid advances in artificial intelligence safely and responsibly in applications that involve specific domain expertise is non-trivial. Our goal in this work was to examine performance of current large language models on health tasks and much improvement is possible with a limit number of training samples. We were not attempting to create a system that would be used in practice. It is well established that large language models can ``hallucinate'' outputs. This is one of the arguments \emph{for} tuning models more specifically and grounding them with data that many more be present in large text corpora. Nonetheless, in health and medical applications, the implications of inaccurate responses are significant.

\section{Conclusion}

Large language models have compelling applications; however, they are known to have limitations when it comes to numerical data. In health, it is vital that insights are based on quantitative measurements. Combining representations encoded into language models with quantitative physiological and behavioral measurements would be a powerful tool in consumer health. We have demonstrated that few shot prompt tuning on a 24B language model can ground time-series data results in large improvements on tasks involving cardiac, metabolic, physical and mental health.

\section{Additional Results of Prompt Engineering \& Zero-Shot}

\begin{table}[h!]
\setlength{\tabcolsep}{3pt}
\caption{\small \textbf{Results.} Comparison of performance between zero-shot LLMs and prompt-tuned LLMs (w/ Context-Inclusive Prompts) across all consumer health tasks.}
\vspace{-8pt}
\label{exp:table:zero_shot}
\small
\begin{center}
\adjustbox{max width=\textwidth}{
\begin{tabular}{ll|c|c|c|a}
\toprule[1.5pt]
& & & \multicolumn{1}{c}{\textbf{Zero-Shot}}  & \multicolumn{1}{c}{\textbf{Prompt Tuned w/ Context}}\\
\cmidrule(lr){4-5} 
\textbf{Topic} & \textbf{Task}  & \textbf{Metric} & 0-Shot & 3-Shot & \% Improvement 
\\ \hline \hline

\multirow{3}{*}{Cardio} & HRs to Average HR & MAE $\downarrow$ (beats/min) & \textbf{2.87} & 6.00 & \textcolor{darkred}{\texttt{-}\textbf{52.1\%}}\\
& IBIs to HR & MAE $\downarrow$ (beats/min) & 15.95 & \textbf{12.3} & \textcolor{darkgreen}{\texttt{+}\textbf{22.9\%}}\\
& IBIs to A.Fib. & Accuracy $\uparrow$ (\%)  & 50.0 & \textbf{85.0} & \textcolor{darkgreen}{\texttt{+}\textbf{70.0\%}}\\ 
& IBIs to Sinus B. & Accuracy $\uparrow$ (\%)  & 40.0 & \textbf{81.0} & \textcolor{darkgreen}{\texttt{+}\textbf{102.5\%}}\\ 
& IBIs to Sinus T. & Accuracy $\uparrow$ (\%)  & 53.0 &  \textbf{65.0} & \textcolor{darkgreen}{\texttt{+}\textbf{22.6\%}}\\ \hline

\multirow{1}{*}{Activity} & IMU Activity & Accuracy $\uparrow$ (\%) & 66.0 & \textbf{62.0} & \textcolor{darkgreen}{\texttt{+}\textbf{6.00\%}}\\ \hline

\multirow{1}{*}{Metabolic} & Calories & MAE $\downarrow$ (calories) & 139 & \textbf{106} & \textcolor{darkgreen}{\texttt{+}\textbf{23.7\%}}\\ \hline

\multirow{2}{*}{MHealth} & Fitbit to Stress & Accuracy $\uparrow$ (\%)  & 60.0 &  \textbf{72.5} & \textcolor{darkgreen}{\texttt{+}\textbf{20.8\%}}\\
& Fitbit to PHQ & Accuracy $\uparrow$ (\%)  & 48.0 & \textbf{49.0} & \textcolor{darkgreen}{\texttt{+}\textbf{2.08\%}}\\

\bottomrule[1.5pt]
\end{tabular}}
\end{center}
\vspace{-0.2cm}
\end{table}

\begin{table}[h!]
\setlength{\tabcolsep}{3pt}
\caption{\small \textbf{Results.} Comparison of performance between Prompt Engineering-ed LLMs and prompt-tuned LLMs (w/ Context-Inclusive Prompts) across all consumer health tasks.}
\vspace{-8pt}
\label{exp:table:prompt_eng}
\small
\begin{center}
\adjustbox{max width=\textwidth}{
\begin{tabular}{ll|c|c|c|a}
\toprule[1.5pt]
& & & \multicolumn{1}{c}{\textbf{Prompt Engineering}}  & \multicolumn{1}{c}{\textbf{Prompt Tuned w/ Context}}\\
\cmidrule(lr){4-5} 
\textbf{Topic} & \textbf{Task}  & \textbf{Metric} & 3-Shot & 3-Shot & \% Improvement 
\\ \hline \hline

\multirow{3}{*}{Cardio} & HRs to Average HR & MAE $\downarrow$ (beats/min) & 27.0 & \textbf{6.00} & \textcolor{darkgreen}{\texttt{+}\textbf{77.8\%}}\\
& IBIs to HR & MAE $\downarrow$ (beats/min) & 33.9 & \textbf{12.3} & \textcolor{darkgreen}{\texttt{+}\textbf{61.1\%}}\\
& IBIs to A.Fib. & Accuracy $\uparrow$ (\%)  & 49.0 & \textbf{85.0} & \textcolor{darkgreen}{\texttt{+}\textbf{73.4\%}}\\ 
& IBIs to Sinus B. & Accuracy $\uparrow$ (\%)  & 56.0 & \textbf{81.0} & \textcolor{darkgreen}{\texttt{+}\textbf{44.6\%}}\\ 
& IBIs to Sinus T. & Accuracy $\uparrow$ (\%)  & 45.0 &  \textbf{65.0} & \textcolor{darkgreen}{\texttt{+}\textbf{44.4\%}}\\ \hline

\multirow{1}{*}{Activity} & IMU Activity & Accuracy $\uparrow$ (\%) & 46.0 & \textbf{62.0} & \textcolor{darkgreen}{\texttt{+}\textbf{34.8\%}}\\ \hline

\multirow{1}{*}{Metabolic} & Calories & MAE $\downarrow$ (calories) & 122 & \textbf{106} & \textcolor{darkgreen}{\texttt{+}\textbf{13.1\%}}\\ \hline

\multirow{2}{*}{MHealth} & Fitbit to Stress & Accuracy $\uparrow$ (\%)  & 68.5 &  \textbf{72.5} & \textcolor{darkgreen}{\texttt{+}\textbf{5.84\%}}\\
& Fitbit to PHQ & Accuracy $\uparrow$ (\%)  & 46.0 & \textbf{49.0} & \textcolor{darkgreen}{\texttt{+}\textbf{6.52\%}}\\

\bottomrule[1.5pt]
\end{tabular}}
\end{center}
\vspace{-0.2cm}
\end{table}

Table \ref{exp:table:zero_shot} illustrates that the context-inclusive prompt-tuned LLM surpasses the zero-shot LLM in performance across all consumer health tasks, with the sole exception being the simplest task of calculating the average heart rate from a list of heart rates. We hypothesize this anomaly could be attributed to overfitting. Further, as depicted in Table \ref{exp:table:prompt_eng}, the context-inclusive prompt-tuned LLM consistently performs better than the prompt-engineered LLM across all tasks, thereby emphasizing the significance of grounding time-series health data through tuning. It's also important to note that the lengthy input sequences of time-series sensor data (e.g., Interbeat Intervals (IBIs) or accelerometer data) can cause issues. The prompt-engineered LLM was unable to produce output for approximately 50\% of the test data in the cardiovascular and activity recognition tasks while prompt-tuned LLM had 0\% failure rate. This outcome further underscores the challenges inherent in managing complex and lengthy data sequences without proper tuning.

\section{Additional Details of Datasets}

\subsection{Cardiovascular Tasks}

For the cardiovascular tasks, we utilize two publicly accessible datasets, MIMIC-III~\cite{johnson2016mimic} and MIT-BIH~\cite{goldberger2000physiobank}, obtained from the Physionet data repository \footnote{www.physionet.org}, maintained by the MIT Computational Physiology Lab.

The MIMIC-III waveform dataset \footnote{https://physionet.org/content/mimiciii/1.4/} represents a comprehensive collection of multimodal biosignal measurements from the ICU stays of over 40,000 patients. Most recordings include multi-lead ECG traces, sampled at 125 Hz, alongside automatically derived metrics such as heart rate, respiration rate, rhythm status classes, and conduction classes. We construct a small subset by identifying six common rhythm status classes and five conduction classes. From each class, we randomly sample 100 15-second lead-II ECG examples, thereby forming a subset with 1,046 examples from 476 unique patients. In this study, we specifically targeted three Interbeat Interval (IBI)-based rhythm status classes, comprising 606 examples and 286 unique patients. These three rhythm status classes, derived from the automatic heart rate measurements, are Normal Sinus, Sinus Bradycardia, and Sinus Tachycardia. We classify a sample as Sinus Bradycardia when the average heart rate is less than 60 beats per minute (bpm) and as Sinus Tachycardia when the average heart rate exceeds 100 bpm.

The MIT-BIH Atrial Fibrillation dataset \footnote{https://physionet.org/content/afdb/1.0.0/} is more narrow in scope and consists of approximately 10-hour-long ECG recordings from 23 patients suffering from paroxysmal atrial fibrillation. Besides the raw 2-lead ECG waveforms, acquired at a 250-Hz sampling rate, the dataset includes rhythm status annotations from clinicians for all the recordings. We select only a small segment of the dataset for the cardiovascular tasks. Firstly, we randomly sample 100 15-second segments from the several hours of lead-I ECG recordings annotated as Normal Sinus Rhythm by clinicians. We then repeat this process for the records labeled as being in Atrial Fibrillation. Both groups include at least one sample from all 23 recordings. To maintain consistency with the MIMIC-III samples, we downsample the waveforms to 125 Hz using the standard resampling algorithm in the \texttt{scipy} Python package.

We employed common post-processing methods to extract the Interbeat Intervals (IBIs) from the ECG traces. IBIs, defined as the time lags between adjacent R-wave peaks in an ECG strip, are crucial for assessing heart rhythm irregularities. For IBI extraction, we utilized the peak detection algorithms provided in the open-source \texttt{ECG-kit} package \footnote{https://github.com/marianux/ecg-kit}.

\subsection{Activity Recognition Task}

We first calcualte the accelerometer magnitude based on the raw data provided in PAMAP2 Physical Activity Monitoring Dataset \cite{reiss2012introducing}. We downsample the data to second-level granularity by taking the mean magnitude of non-overlapping 1-second windows. For each prompt, a 5-second window of accelerometer magnitudes is included. For our training, validation, and testing stages, we use the first 100 5-second windows, across both running and talking tasks (the first 25 for training, the second 25 for validation and the last 50 for testing). To be more specific, walking windows are derived from subjects 101 and 102, whereas running windows are extracted from subjects 101, 102, 104, and 105. The selection of data from different subjects for walking and running tasks is due to the varying number of windows available for each activity per subject. This discrepancy highlights the reason why these activities originate from different subjects.

\subsection{Mental Health Tasks}
For the mental health tasks we used a dataset of wearable data associated with survey instruments collected via a four-week intensive longitudinal study.  Participants who chose to enroll in this study were directed to a consent form administered via an in-app onboarding flow that described the data that would be collected and how it would be used to advance the goals of the study. Prospective study participants agreed to an E-consent, which provided explanations for the data they are sharing and how they are used and a long form consent which was a full IRB approved consent form detailing the data use and rights. This was designed to ensure that participants had the best possible comprehension. The consent form was associated with a unique participant ID that is stored securely. The purpose of this ID is to be able to link the consent form to the data collected, in order to confirm that data was collected with consent if needed. As mobile sensing is nonintrusive participants may easily forget they are participating in the study, the participants received daily notifications reminding them that they are enrolled in the study. Participants were informed that if they choose to withdraw, they would have the option to remove or retain all or part of their data. In addition, even if they do not withdraw from the study, they could choose at any time to have any portion of their data omitted, giving the dates and times between which they want their data to be removed.
In our analysis for this paper we randomly sampled 150 subjects' data for the purposes of daily stress EMA and PHQ score estimation.  The terms of the IRB mean that these data cannot be made publicly available. However, more details of the study can be found in the published study protocol [REMOVE FOR ANONYMIZATION].

\bibliographystyle{abbrv}
\bibliography{references}

\end{document}